\documentclass[10pt,twocolumn,letterpaper]{article}

\usepackage[utf8]{inputenc} 
\usepackage{iccv}
\usepackage{times}
\usepackage{epsfig}
\usepackage{graphicx}
\usepackage{amsmath}
\usepackage{amssymb}
\usepackage{booktabs}
\usepackage{subcaption}
\usepackage{makecell}
\usepackage{soul}
\usepackage{mathtools}
\usepackage[pagebackref=true,breaklinks=true,letterpaper=true,colorlinks,bookmarks=false]{hyperref}

\hypersetup{
	plainpages=false,
	colorlinks=true,              %
	linkcolor=blue,               %
	anchorcolor=blue,             %
	citecolor=blue,               %
	filecolor=blue,               %
	urlcolor=blue,                %
	pdfview=FitH,                 %
	pdfstartview=FitH,            %
	pdfpagelayout=SinglePage      %
}

\iccvfinalcopy 

\def\iccvPaperID{45} 

\newcommand{\bv}{\mathbf{v}}
\newcommand{\bc}{\mathbf{c}}

\makeatletter
\def\thanks#1{\protected@xdef\@thanks{\@thanks
        \protect\footnotetext{#1}}}
\makeatother

\begin{document}

\title{HowTo100M: Learning a Text-Video Embedding by \\ Watching Hundred Million Narrated Video Clips}

\author{Antoine Miech$^{1,2*}$ \qquad Dimitri Zhukov$^{1,2*}$\thanks{$^*$Equal contribution.}  \qquad Jean-Baptiste Alayrac$^{2+}$\thanks{$^{+}$Now at DeepMind.}     \\
Makarand Tapaswi$^{2}$ \qquad Ivan Laptev$^{1,2}$ \qquad Josef Sivic$^{1,2,3}$\\
$^1$École Normale Supérieure\thanks{$^1$Département d’informatique de l’ENS, École normale supérieure, CNRS, PSL Research University, 75005 Paris, France.} \quad \quad \quad $^2$Inria\quad \quad \quad $^3$CIIRC, CTU\thanks{$^3$Czech Institute of Informatics, Robotics and Cybernetics at the Czech Technical University in Prague.}\\
\url{https://www.di.ens.fr/willow/research/howto100m}\\
}

\maketitle

\begin{abstract}

Learning text-video embeddings usually requires a dataset of video clips with manually provided captions.
However, such datasets are expensive and time consuming to create and therefore difficult to obtain on a large scale.
In this work, we propose instead to learn such embeddings from video data with readily available natural language annotations in the form of automatically transcribed narrations.
The contributions of this work are three-fold.
First, we introduce \emph{HowTo100M}: a large-scale dataset of 136 million video clips sourced from 1.22M narrated instructional web videos depicting humans performing and describing over 23k different visual tasks.
Our data collection procedure is fast, scalable and does not require any additional manual annotation.
Second, we demonstrate that a text-video embedding trained on this data leads to state-of-the-art results for text-to-video retrieval and action localization on instructional video datasets such as YouCook2 or CrossTask. 
Finally, we show that this embedding transfers well to other domains: fine-tuning on generic Youtube videos (MSR-VTT dataset) and movies (LSMDC dataset) outperforms models trained on these datasets alone.
Our dataset, code and models are publicly available~\cite{iccv2019howto100m}.

\end{abstract}

\begin{figure}[t]
\centering
\includegraphics[width=\columnwidth]{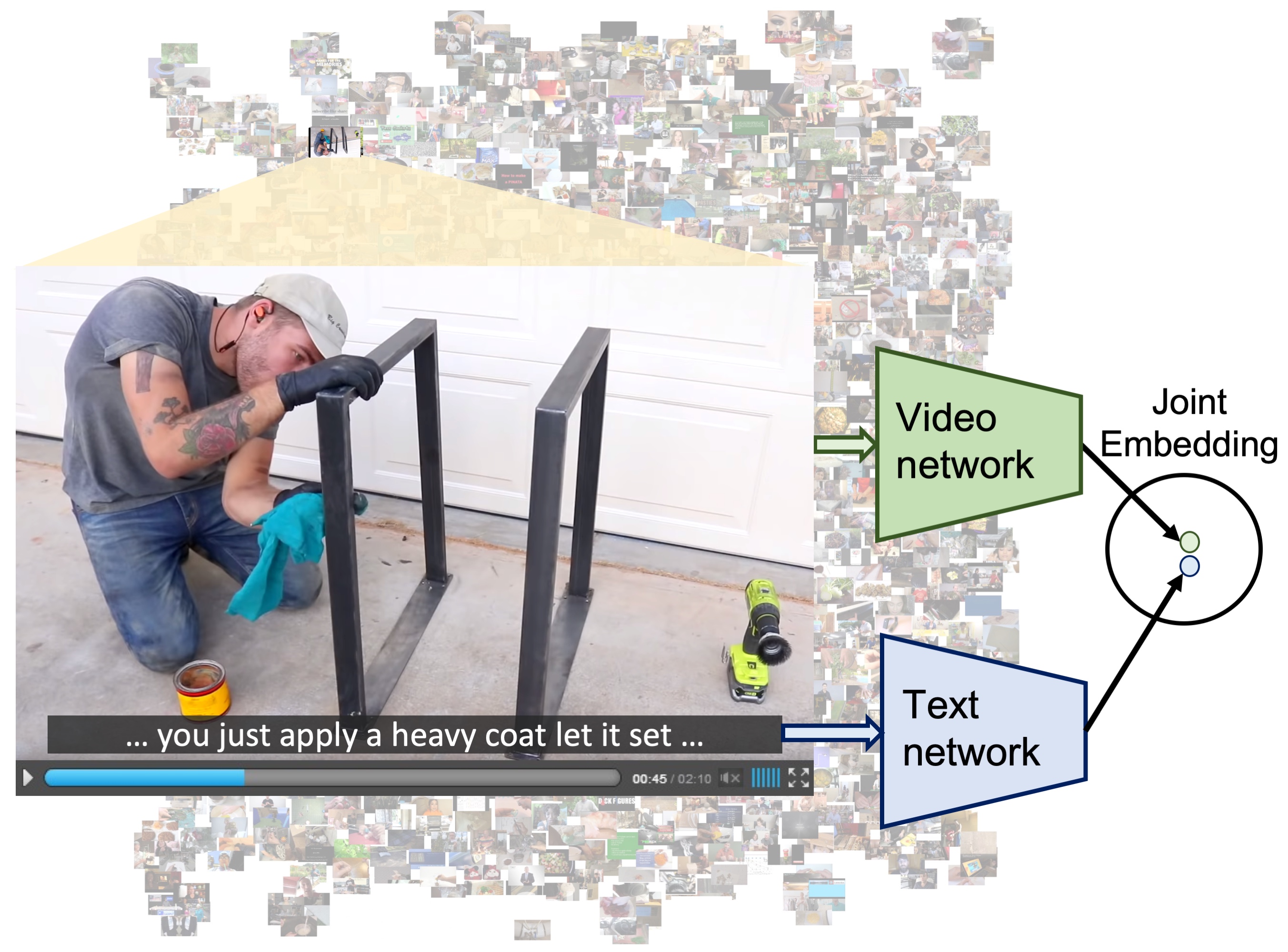}
\vspace{-6mm}
\caption{\small We learn a joint text-video embedding by watching millions of narrated video clips of people performing diverse visual tasks.  The learned embedding transfers well to other instructional and non-instructional text-video datasets.}
\vspace{-4mm}
\label{fig:teaser}
\end{figure}

\section{Introduction}
\label{section:intro}


Communicating about the visual world using language is a key ability of humans as intelligent beings.
A three year old child can manipulate objects, observe its own actions and describe them to others using language;
while adults can learn new skills by reading books or watching videos.
This interplay between video and language extends naturally to artificial agents that need to understand the visual world and communicate about it with people.
Examples of tasks that still represent a significant challenge for current artificial systems include text-to-video retrieval~\cite{klein15associating,miech18learning,wang2018learning,wang2016learning,yu18joint}, text-based action or event localization~\cite{hendricks17localizing}, video captioning~\cite{pan16hierarchical,yu2016video}, and video question answering~\cite{tapaswi16movieqa,yu18joint}.
Yet, progress on these problems is important for a host of applications from searching video archives to human-robot communication.

A common approach to model visual concepts
described with language is to learn a mapping of text and video into a shared embedding space, where related text fragments and video clips are close to each other~\cite{hendricks17localizing,miech18learning,pan16jointly,plummer2017enhancing,xu2015jointly}.
Learning a good representation often requires a large set of paired video clips and text captions.
In fact, given the huge variability of video scenes and their textual descriptions, learning a generic embedding space may require millions of paired video clips and text captions.
However, existing datasets (\eg~MSR-VTT~\cite{xu16msrvtt}, DiDeMo~\cite{hendricks17localizing}, EPIC-KITCHENS~\cite{damen2018scaling}), are on the scale of tens to hundreds of thousands of such pairs that have been annotated manually.
Manual collection of such datasets is expensive and hard to scale. It is also subjective since video annotation can often be an ill-defined task with low annotator consistency~\cite{xu16msrvtt}.

In this work, we explore a different source of supervision to obtain paired video clips and text captions for learning joint representations of video and language.
We observe that \emph{narrated instructional videos} are available in large quantities (\eg~on YouTube) and provide a large amount of visual and language data.
In particular, instructional videos~\cite{alayrac16unsupervised,malmaud15what,zhukov2019crosstask} often contain narration with an explicit intention of explaining the visual content on screen. 
To leverage this rich source of data, we collect a new large-scale dataset containing 136 million video clips sourced from 1.22 million narrated instructional videos depicting humans performing more than 23,000 different tasks.
Each clip is paired with a text annotation in the form of an automatically transcribed narration.


{\bf \noindent Contributions.}
The contributions of this work are three-fold.
First, we collect a new dataset of close-captioned video clips, \emph{HowTo100M}, that is orders of magnitude larger than any other existing video-text datasets (Section~\ref{section:dataset}).
Second, we show that such data can be used to learn powerful video-language representations.
Our model (Section~\ref{section:model}), trained on HowTo100M, sets a new state-of-the-art for text-based action localization and text-to-video retrieval on existing datasets of instructional videos, YouCook2~\cite{youcook2} and CrossTask~\cite{zhukov2019crosstask}.
Finally, we explore the ability of models trained on our data to transfer to non-instructional videos.
In particular, we demonstrate that models pretrained on HowTo100M can be successfully transferred by fine tuning
on the MSR-VTT dataset (generic Youtube videos) and the LSMDC dataset (movies).

\section{Related work}
\label{section:related}

A significant number of computer vision applications rely on a joint understanding of visual and textual cues.
These applications include automatic image and video captioning~\cite{johnson16densecap,pan16hierarchical,you16image,yu2016video}, visual question answering~\cite{fukui16multimodal,malinowski15ask,tapaswi16movieqa,yu18joint}, visual content retrieval based on textual queries~\cite{miech18learning,wang2019learning,yu18joint}, temporal localization of events in videos using natural language~\cite{hendricks17localizing,krishna2017dense} or video summarization with natural language~\cite{plummer2017enhancing}.

\vspace{1mm}
\noindent
\textbf{Vision, language and speech.}
A common approach to model vision and language is learning a joint embedding space where visual and textual cues are adjacent if and only if they are semantically similar~\cite{chowdhury2018webly,dong19dual,gong14multi,gong14improving,klein15associating,miech18learning,mithun2018learning,pan16jointly,plummer2017enhancing,xu2015jointly,wang2018learning,wang2016learning,wu2017sampling}.
Most of these works rely on medium scale well annotated datasets in which descriptive captions are collected for each video clip.
This process is costly as it requires considerable human annotation effort making these datasets hard to scale (see Table~\ref{table:dataset-comparison}).
In this work, we train a joint video and language model \textit{without a single manually annotated video description} by leveraging automatically transcribed narrated videos.
Using the spoken text from narrated videos to supervise vision models has seen some recent interest~\cite{alayrac16unsupervised,chen17discover,harwath18jointly, malmaud15what,sanabria18how2,yu14instructional}.
Harwath~\etal~\cite{harwath18jointly} utilize the raw speech waveform to supervise the visual model, however, their method does not scale as annotators were paid to record audio descriptions for thousands of images.
Chen~\etal~\cite{chen17discover} use subtitles from documentaries to automatically obtain object labels, but their focus is on learning object detectors rather than text-video embeddings and their dataset contains only 9 documentary movies, compared to about 15 years of video content considered in this work. 

\begin{table}[t]
  \setlength{\tabcolsep}{3pt}
    \centering
    \scalebox{0.75}{
    \begin{tabular}{@{}lrrrrrr@{}}
      \toprule
      Dataset & Clips & Captions & Videos & Duration   & Source  & Year \\
      \midrule
      Charades~\cite{sigurdsson2016hollywood} & 10k & 16k & 10,000 & 82h  & Home  & 2016 \\
       MSR-VTT~\cite{xu16msrvtt} & 10k & 200k & 7,180 & 40h  & Youtube  & 2016 \\
       YouCook2~\cite{youcook2} & 14k & 14k & 2,000 & 176h  & Youtube  & 2018   \\
       EPIC-KITCHENS~\cite{damen2018scaling} & 40k & 40k & 432 & 55h  & Home  & 2018   \\
      DiDeMo~\cite{hendricks17localizing} & 27k & 41k & 10,464 & 87h & Flickr  & 2017  \\
         M-VAD~\cite{torabi15using} & 49k & 56k & 92 & 84h  & Movies  & 2015 \\
        MPII-MD~\cite{rohrbach15dataset} & 69k & 68k & 94 & 41h  & Movies  & 2015 \\
       ANet Captions~\cite{krishna2017dense} & 100k & 100k & 20,000 & 849h  & Youtube  & 2017  \\
      TGIF~\cite{tgif-cvpr2016} & 102k & 126k & 102,068 & 103h  & Tumblr  & 2016 \\
        LSMDC~\cite{rohrbach17movie} & 128k & 128k & 200 & 150h  & Movies  & 2017 \\
        How2~\cite{sanabria18how2}  & 185k & 185k  & 13,168 & 298h& Youtube  & 2018  \\
      \textbf{HowTo100M} & \textbf{136M} & \textbf{136M}  & \textbf{1.221M} & \textbf{134,472h} & Youtube  & 2019 \\
      \bottomrule
    \end{tabular}
    }
\vspace{-2mm}
\caption{Comparison of existing video description datasets. The size of our new HowTo100M dataset bypasses the size of largest available datasets by three orders of magnitude. M~denotes million while k denotes thousand.}
\vspace{-4mm}
      \label{table:dataset-comparison}
\end{table}

\vspace{1mm}
\noindent
\textbf{Learning from instructional videos}. 
Instructional videos are rising in popularity in the context of learning steps of complex tasks~\cite{alayrac16unsupervised,feifei2016connectionist,
richard17weakly,richard18actionsets,sener18unsupervised,zhukov2019crosstask},
 visual-linguistic reference resolution~\cite{huang17unsupervised,huang18finding}, action segmentation in long untrimmed videos~\cite{zhou18towards} and joint learning of object states and actions~\cite{alayrac16objectstates}.
Related to our work, \cite{alayrac16unsupervised,malmaud15what,yu14instructional} also consider automatically generated transcription of narrated instructional videos as a source of supervision. 
However as opposed to our work, these works typically extract from transcriptions only a small number of predefined labels.

\begin{figure*}[t]
\centering
\includegraphics[width=\linewidth]{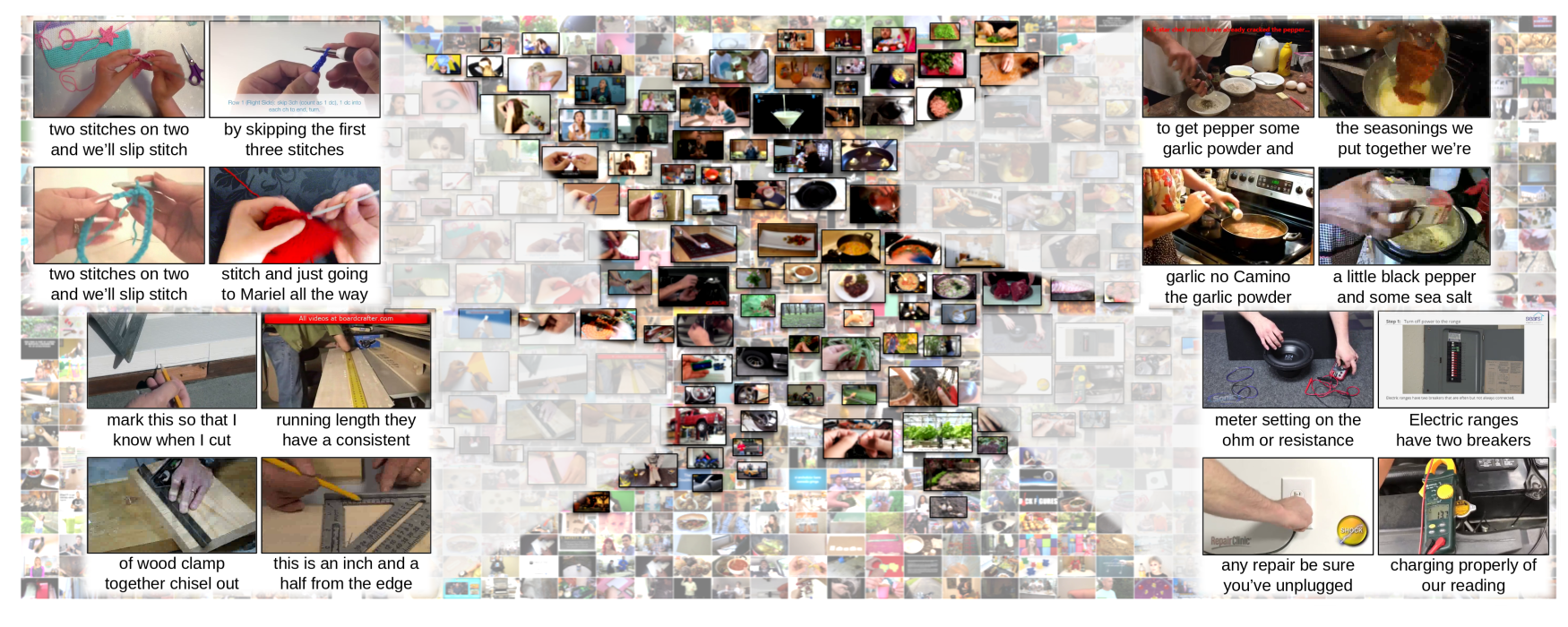}
\vspace{-8mm}
\caption{\small Examples of clip-caption pairs retrieved with the help of our joint embedding. Pairs are selected based on the similarity between visual appearance and corresponding narration, while they are arranged based on linguistic similarity across pairs. Examples are taken from 4 distinct clusters, corresponding to \textit{Knitting}, \textit{Woodwork/Measuring}, \textit{Cooking/Seasoning} and \textit{Electric maintenance}.}
\vspace{-3mm}
\label{fig:pair-examples}
\end{figure*}

Numerous datasets of web instructional videos were proposed over the past years~\cite{alayrac16unsupervised,malmaud15what,
sanabria18how2,Sener_2015_ICCV,
tang2019coin,youcook2,zhukov2019crosstask}.
Among the first to harvest instructional videos, Sener~\etal~\cite{Sener_2015_ICCV} use WikiHow, an encyclopedia of \emph{how to} articles, to collect 17 popular physical tasks, and obtain videos by querying these tasks on YouTube.
In a similar vein, COIN~\cite{tang2019coin} and CrossTask~\cite{zhukov2019crosstask} datasets are collected by first searching for tasks on WikiHow and then videos for each task on YouTube.
We use the same approach for collecting HowTo100M.
The main distinction between our dataset and previous efforts is the unprecedented scale both in terms of variety (more than 23,000 tasks from 12 different domains) and size (136 million clips sourced from 1.2 million instructional videos).

\vspace{1mm}
\noindent
\textbf{Large scale data for model pretraining.}
The use of large scale and potentially noisy data from the web is an exciting prospect to pretrain language and vision models.
In natural language processing, BERT~\cite{bert18}, GPT~\cite{gpt18openai}, and GPT-2~\cite{radford19LM} are examples of language models trained on large-scale data that achieve state-of-the-art for many tasks.
In fact, training GPT-2 on WebText~\cite{radford19LM} a dataset of 40GB of text from Reddit achieves state-of-the-art even in \emph{zero-shot} settings.
In vision, \cite{limitsWeakly18mahajan,sun17JFT} explore the use of image metadata such as Instagram hashtags to pretrain image classifiers.

We are inspired by these works and focus our efforts on learning a strong embedding for joint understanding of video and language.
We demonstrate that our video-language embedding learned from millions of YouTube videos not only outperforms previous work on tasks related to instructional videos without fine-tuning, but also generalizes well to non-instructional videos with some fine-tuning.
We release our dataset, feature extraction pipeline, and model parameters as a resource that the video and language community can build on.



\section{The HowTo100M dataset}
\label{section:dataset}

We collect a new dataset of narrated videos with an emphasis on instructional videos where content creators teach complex tasks.
This ensures that most narrations describe the observed visual content.
HowTo100M features 1.22 million videos from YouTube, with activities from domains such as cooking, hand crafting, personal care, gardening, \etc.
Each video is associated with a narration available as subtitles that are either written manually or are the output of an Automatic Speech Recognition (ASR) system.


\subsection{Data collection}
\label{data_collection}

\noindent
\textbf{Visual tasks.}
With an aim to obtain instructional videos that describe how to perform certain activities, we first start by acquiring a large list of activities using \emph{WikiHow}\footnote{\url{https://www.wikihow.com}} -- an online resource that contains 120,000 articles on \textit{How to ...} for a variety of domains ranging from cooking to human relationships structured in a hierarchy.
We are primarily interested in ``visual tasks'' that involve some interaction with the physical world (\eg~\emph{Making peanut butter}, \emph{Pruning a tree}) as compared to others that are more abstract (\eg~\emph{Ending a toxic relationship}, \emph{Choosing a gift}).
To obtain predominantly visual tasks, we limit them to one of 12 categories (listed in Table~\ref{table:dataset-categories}).
We exclude categories such as \textit{Relationships} and \textit{Finance and Business}, that may be more abstract.

We further refine the set of tasks, by filtering them in a semi-automatic way.
In particular, we restrict the primary verb to physical actions, such as \textit{make}, \textit{build} and \textit{change}, and discard non-physical verbs, such as \textit{be}, \textit{accept} and \textit{feel}.
This procedure yields 23,611 visual tasks in total.

\vspace{1mm}
\noindent
\textbf{Instructional videos.}
We search for YouTube videos related to the task by forming a query with \emph{how to} preceding the task name (\eg~\textit{how to paint furniture}).
We choose videos that have English subtitles - either uploaded manually, generated automatically by YouTube ASR, or generated automatically after translation from a different language by YouTube API.

We improve the quality and consistency of the dataset, by adopting the following criteria.
We restrict to the top 200 search results, as the latter ones may not be related to the query task.
Videos with less than 100 views are removed as they are often of poor quality or are amateurish.
We also ignore videos that have less than 100 words as that may be insufficient text to learn a good video-language embedding.
Finally, we remove videos longer than 2,000 seconds.

As some videos may appear in several tasks, we de-duplicate videos based on YouTube IDs.
However, note that the dataset may still contain duplicates if a video was uploaded several times or edited and re-uploaded.
Nevertheless, this is not a concern at our scale.


\subsection{Paired video clips and captions}

Subtitles are often organized as a list of text chunks (lines), and need not form complete sentences.
Each line is associated with a time interval in the video, typically the duration in which the line is uttered.
We select each line of the subtitles as a caption, and pair it with the video clip from the time interval corresponding to the line.
We show some examples from our clip-caption pairs in Figure~\ref{fig:pair-examples}.

Different from other datasets with clip-caption pairs (\eg~MSR-VTT), our captions are \emph{not} manually annotated, but automatically obtained through the narration.
Thus, they can be thought of as \emph{weakly paired}.
Typical examples of incoherence include the content producer asking viewers to subscribe to their channel, talking about something unrelated to the video, or describing something before or after it happens.
Furthermore, our captions are often incomplete, lack punctuation, or are grammatically incorrect sentences, as they come from continuous narration and often ASR.
We have manually inspected 400 randomly sampled clip-caption pairs and found that in 51 \%, at least one object or action mention in the caption is visually seen in the video clip.

\begin{table}[t]
  \setlength{\tabcolsep}{3pt}
    \centering
        \scalebox{0.9}{
    \begin{tabular}{@{}lrrr@{}}
      \toprule
      Category & Tasks & Videos & Clips \\
      \midrule
		Food and Entertaining & 11504 & 497k & 54.4M \\
		Home and Garden & 5068 & 270k & 29.5M \\
		Hobbies and Crafts & 4273 & 251k & 29.8M \\
		Cars \& Other Vehicles & 810 & 68k & 7.8M \\
		Pets and Animals & 552 & 31k & 3.5M \\
		Holidays and Traditions & 411 & 27k & 3.0M \\
		Personal Care and Style & 181 & 16k & 1.6M \\
		Sports and Fitness & 205 & 16k & 2.0M \\
		Health & 172 &	15k & 1.7M \\
		Education and Communications & 239 & 15k & 1.6M \\
		Arts and Entertainment & 138 & 10k & 1.2M \\
		Computers and Electronics & 58 & 5k & 0.6M \\
      \midrule
      Total & 23.6k & 1.22M & 136.6M \\
      \bottomrule
    \end{tabular}
    }
 \vspace{-1mm}
 \caption{Number of tasks, videos and clips within each category.}
      \label{table:dataset-categories}
\end{table}

\vspace{1mm}
\noindent
\textbf{Statistics.}
The initial set of visual tasks are obtained by focusing on 12 WikiHow categories.
Table~\ref{table:dataset-categories} shows the number of collected WikiHow tasks and corresponding videos and clips per category.
In Appendix A~\cite{miech19howto100m}, we show the first two levels of the WikiHow hierarchy: the twelve categories and their subcategories along with the number of chosen tasks and corresponding videos in our dataset.
We compare the sizes of existing clip-caption paired datasets in Table~\ref{table:dataset-comparison}. 
HowTo100M is several orders of magnitude larger than existing datasets and contains an unprecedented duration (15 years) of video data.
However, unlike previous datasets, HowTo100M does not have clean annotated captions.
As the videos contain complex activities, they are relatively long with an average duration of 6.5 minutes.
On average, a video produces 110 clip-caption pairs, with an average duration of 4 seconds per clip and 4 words (after excluding stop-words) per caption.
For more details, we show in Appendix A~\cite{miech19howto100m} the distribution of nouns and verbs.
Our data collection procedure assumes that searching with \textit{How to} queries on YouTube would result in mostly instructional videos.
We verify this by randomly selecting 100 videos and labeling their type.
71\% of the videos are found to be instructional, 12\% are vlogs, and another 7\% are product reviews or advertisements.
Note that vlogs, reviews and ads may also contain correspondences between visual content and narration.
In particular, we noticed that objects shown on screen are often mentioned in narration.
We do not discard such non-instructional videos, as they may still be useful for the learning the joint embedding.

\section{Text-video joint embedding model}
\label{section:model}

We now present our model to learn a joint text-video embedding from the automatically paired video clips and captions in our dataset.
%
More formally, we are given a set of $n$ video clips and associated captions $\{(V_i, C_i)\}_{i=1}^n$.
We denote by $\bv \in \mathbb{R}^{d_v}$ and $\bc \in \mathbb{R}^{d_c}$ the $d_v$ and $d_c$ dimensional feature representation of a video clip $V$ and caption $C$, respectively.
Given this, our goal is to learn two mapping functions: $f:\mathbb{R}^{d_v}\to\mathbb{R}^{d}$ and $g:\mathbb{R}^{d_c}\to\mathbb{R}^{d}$
that respectively embed video and caption features into a common $d$-dimensional space, such that the cosine similarity
\begin{equation}
\label{eq:sim}
 s(V, C) = \frac{\langle f(\bv), g(\bc) \rangle}{\|f(\bv)\|_2 \|g(\bc)\|_2}
\end{equation}
is high when caption $C$ describes the video clip $V$, and low otherwise.

In this work, we use the class of non-linear embedding functions used in~\cite{miech18learning}, which are given by:
\begin{align}
\label{eq:f}
 f(\bv) &= (W^v_1 \bv + b^v_1) \circ \sigma(W^v_2 (W^v_1 \bv + b^v_1) + b^v_2)\\
\label{eq:g}
\text{and } g(\bc) &= (W^c_1 \bc + b^c_1) \circ \sigma(W^c_2 (W^c_1 \bc + b^c_1) + b^c_2),
\end{align}
where $W^v_1 \in \mathbb{R}^{d \times d_v}$, $W^c_1 \in \mathbb{R}^{d \times d_c}$, $W^v_2, W^c_2 \in \mathbb{R}^{d \times d}$, $b^v_1, b^c_1, b^v_2, b^c_2 \in \mathbb{R}^{d}$ are learnable parameters,
$\sigma$ is an element-wise sigmoid activation and
$\circ$ is the element-wise multiplication (Hadamard product).
In practice, $d_v=4,096$, $d_c=4,096$ and $d=4,096$ resulting in a model composed of 67M parameters. 
Note that the first term on the right-hand side in Equations \eqref{eq:f} and \eqref{eq:g} is a linear fully-connected layer and the second term corresponds to a context gating function~\cite{miech17learnable} with an output ranging between 0 and 1, which role is to modulate the output of the linear layer.
As a result, this embedding function can model non-linear multiplicative interactions between the dimensions of the input feature vector which has proven effective in other text-video embedding applications~\cite{miech18learning}.

{\bf \noindent Loss.}
We train our embedding model using the max-margin ranking loss~\cite{karpathy14deepfragment,miech18learning,wang2018learning,wang2016learning,yu16videocaptioning}.
At each iteration of our training algorithm, we sample a mini-batch $\mathcal{B} = \{ i_1, ..., i_b\} \subset \{1,\dots,n\}$ of caption-clip training pairs $(V_{i}, C_{i})_{i \in \mathcal{B}}$, and update the model parameters with a gradient step of the following loss:
\label{loss}
\[\sum_{i \in \mathcal{B}} \smashoperator[r]{\sum_{j \in \mathcal{N}(i)}} \max(0, \delta + s_{i,j} - s_{i,i}) + \max(0, \delta + s_{j,i} - s_{i,i}),\]
where $s_{i,j} = s(V_{i}, C_{j})$ is the similarity score~\eqref{eq:sim} between video clip $V_i$ and caption $C_j$, $\mathcal{N}(i)$ is a set of negative pairs for caption-clip $i$ and $\delta$ is the margin.
The first term in Equation~\eqref{loss} corresponds to the ranking loss when sampling a negative caption, while the second term corresponds to sampling a negative video clip.
We fix $\delta = 0.1$ in practice.
Our model parameters are updated using Adam~\cite{kingma15adam} with a learning rate of $10^{-4}$.
Implementation details of the loss are provided in Appendix B~\cite{miech19howto100m}.

{\bf \noindent Sampling strategy.}
Similar to~\cite{hendricks17localizing}, we apply an intra-video negative sampling strategy to define $\mathcal{N}(i)$.
We show in Section~\ref{subsec:sampling} that this approach is \emph{critical} for good performance.
More precisely, half of our negative pairs $\{(V_i, C_j) :i \neq j\}$, are selected such that the video clip $V_i$ and the caption $C_j$ belong to the same original YouTube video (as $(V_i, C_i)$), while the other half are sampled from other YouTube videos.
We apply intra-negative sampling to ensure that the learned embedding focuses on relevant aspects of the video clip (\eg~the hands of the person showing how to knead dough) rather than irrelevant background features (\eg~the kitchen).
In Appendix C~\cite{miech19howto100m}, we also provide an empirical analysis of the positive pair sampling strategy.
We show that even though the training data is noisy, our attempts to automatically select correct positive pairs during training did not yield improvements so far. We think this could be attributed to the fact our model is shallow and is trained on a large amount of data.

{\bf \noindent Clip and caption representation.}
The clip feature $\bv$ consists of temporally max-pooled pre-extracted CNN features.
The caption feature $\bc$ is the output of a shallow 1D-CNN on top of pre-computed word embeddings.
More details are given in Section~\ref{implementation_details}.

\section{Experiments}

In this section, we demonstrate that a strong joint representation for video and text can be learned from our unlabeled HowTo100M dataset.
We provide experimental results for a variety of domains ranging from instructional videos in CrossTask~\cite{zhukov2019crosstask},
cooking videos in YouCook2~\cite{youcook2}, generic YouTube videos in MSR-VTT~\cite{xu16msrvtt} to movie video clips in LSMDC~\cite{rohrbach17movie}.
Specifically, we evaluate our learned embedding on the tasks of localizing steps in instructional videos of CrossTask~\cite{zhukov2019crosstask}
and text-based video retrieval on YouCook2~\cite{youcook2}, MSR-VTT~\cite{xu16msrvtt} and LSMDC~\cite{rohrbach17movie} datasets.
 
Our \emph{key} findings are the following:
\textbf{(i)}~For instructional video datasets, such as CrossTask~\cite{zhukov2019crosstask} and YouCook2~\cite{youcook2},
our off-the-shelf embedding trained on HowTo100M significantly outperforms state-of-the-art models trained on much smaller and manually-annotated datasets.
\textbf{(ii)}~On generic YouTube videos (MSR-VTT~\cite{xu16msrvtt}), our HowTo100M embedding provides competitive retrieval performance compared to state-of-the-art methods trained on MSR-VTT.
Moreover, we show that fine-tuning our pre-trained embedding model on just a fifth of annotated videos from MSR-VTT outperforms state-of-the-art.
\textbf{(iii)}~We show that fine-tuning our embedding on LSMDC enables generalization to movie videos and scripts despite the large domain gap. 
\textbf{(iv)}~Finally, we demonstrate the importance of scale in HowTo100M to learn better joint video-text embeddings.

\begin{table}[t]
  \setlength{\tabcolsep}{3pt}
    \centering  
        \resizebox{\linewidth}{!}{
      \begin{tabular}{@{}lccccc@{}}
      \toprule
      Negative sampling & M (R@10) & L (R@10) & Y (R@10)  & C (AVG Recall)  \\
      \midrule
      No intra-negative  & \textbf{30.1} & 12.3 & 18.1 & 25.7 \\
      With intra-negative  & 29.6 & \textbf{14.0} & \textbf{24.8} & \textbf{33.6} \\
      \bottomrule
    \end{tabular}
    }
\vspace{-2mm}
 \caption{Impact of intra-video negative pairs during training. M: MSR-VTT, L: LSMDC, Y: YouCook2, C: CrossTask.}
 \vspace{-4mm}
      \label{table:sampling-experiment}
\end{table}

\subsection{Implementation details}
\label{implementation_details}
{\bf \noindent Video features.} 
We extract frame-level and video-level features with pre-trained 2D and 3D CNNs.
2D features are extracted with the ImageNet pre-trained Resnet-152~\cite{he16resnet} at the rate of one frame per second.
3D features are extracted with the Kinetics~\cite{carreira2017quovadis} pre-trained ResNeXt-101 16-frames model~\cite{hara183dcnns} to obtain 1.5 features per second.
We aggregate features from longer video clips by the temporal max-pooling and concatenate 2D and 3D features to form a single 4096 dimensional vector for each video clip.

{\bf \noindent Text pre-processing.} 
We preprocess transcribed video narrations by discarding common English stop-words. 
For the word representations, we use the GoogleNews pre-trained word2vec embedding model~\cite{mikolov13efficient}.

{\bf \noindent Training time.} 
Once the video and text features are extracted,
training our embedding model on the full HowTo100M dataset is relatively fast and takes less than three days on a single Tesla P100 GPU.

\subsection{Datasets and evaluation setups}
\label{dataset_evaluation}

\paragraph{Action step localization.}
We evaluate localization of action steps in instructional videos on the recent CrossTask dataset~\cite{zhukov2019crosstask}.
CrossTask includes 18 tasks and 2.7k instructional videos with manually annotated action segments. 
Each video may contain multiple segments, corresponding to different actions. 
It also provides an ordered list of action steps with short natural language descriptions for each task. 
We apply our model trained only on HowTo100M to the problem of step localization by computing similarity between every frame in the video and the action label names of CrossTask.
In order to compare to \cite{zhukov2019crosstask}, we follow a similar inference procedure.
We use the same recall metric as in \cite{zhukov2019crosstask}, which is defined by the number of step assignments that fall into the correct ground truth interval, divided by the total number of steps in the video.
Videos from the test set of CrossTask are removed from the HowTo100M training set to ensure that they are not observed at training time.

\noindent
\textbf{Text-based video retrieval.}
We also evaluate our learned embedding on the task of video clip retrieval using natural language queries.
Given a textual description, the goal is to retrieve representative video clips from a large pool of videos.
We evaluate our learned embedding using the standard recall metrics R@1, R@5, R@10 and the median rank (Median R).
We provide experimental results for the following domain-specific video description datasets.

{\bf  YouCook2}~\cite{youcook2} is a cooking video dataset collected from YouTube.
It features 89 different recipes and 14k video clips all annotated with textual descriptions collected from paid human workers.
Since no descriptions are provided for the test set clips, we evaluate YouCook2 clip retrieval task on the validation clips (3.5k in total). 
Note that we have taken care to remove the few validation YouCook2 videos that are also present in HowTo100M.
 
{\bf  MSR-VTT}~\cite{xu16msrvtt} is a dataset of generic videos collected from 257 popular video queries depicting 20 categories (including music, sports, movie, \etc) from YouTube.
It contains 200k unique video clip-caption pairs, all annotated by paid human workers.
We evaluate our model on the MSR-VTT clip retrieval test set used in~\cite{yu18joint} as performance of several other methods is reported on it.

{\bf  LSMDC}~\cite{rohrbach17movie} is a dataset of movie clips.
It features 101k unique video clip-caption pairs.
All clips are associated with a description that either comes from the movie script or the audio description.
We evaluate our model on the official LSMDC test set\footnote{\url{https://sites.google.com/site/describingmovies/lsmdc-2016/movieretrieval}} that contains 1000 video-caption pairs.

\subsection{Study of negative pair sampling strategy}
\label{subsec:sampling}
We first study the effect of alternative strategies for sampling negative caption-video clip pairs when training our embedding.
Table~\ref{table:sampling-experiment} shows that using negatives from the same video (intra-negatives) is beneficial as compared to randomly sampling them from other YouTube videos.
The improvement is particularly significant on YouCook2 and CrossTask which are more fine-grained datasets than MSR-VTT and LSMDC.
For the rest of the paper, we report numbers using our model trained with the intra-negative sampling strategy.

\subsection{Scale matters}
\begin{figure}[t]
  \begin{center}
    \includegraphics[width=\columnwidth]{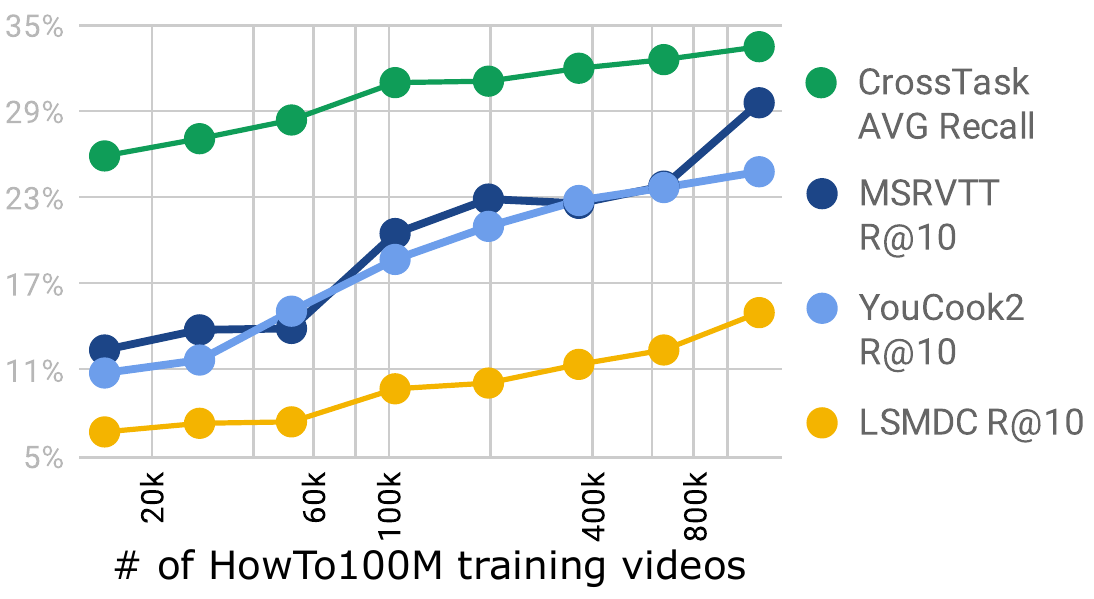}
    \vspace{-.8cm}\\
\end{center}
  \caption{Retrieval and step localization results when varying the training size of our HowTo100M dataset.\vspace{-.3cm}}
  \label{fig:dataset_size}
\end{figure}

A natural question is whether the large scale of our dataset is truly required to achieve high performance.
To answer this, we train our embedding model on smaller subsets of our dataset.
These smaller subsets of HowTo100M are created by gradually decreasing the allowed Youtube search rank (see the paragraph on data collection in Section~\ref{data_collection} for more details) for training videos.
We experiment with the following rank thresholds: top 2 (15k videos), top 3 (28k videos), top 5 (52k videos), top 10 (104k videos), top 20 (197k videos), top 40 (364k videos), top 80 (648k videos) and top 200 (entire HowTo100M dataset).
This process ensures that we subsample training videos that are more likely to be relevant to the queried task as we reduce the size of the training dataset.
Figure~\ref{fig:dataset_size} shows average recall on CrossTask and the R@10 clip retrieval results on LSMDC, MSR-VTT and YouCook2 when varying the size of the training dataset.
There is a clear improvement over all evaluated tasks with the gradual increase in the amount of training data.
Interestingly, we do not observe any saturation, hence we can expect further improvements by collecting even more readily-available and unlabeled video data.
 
\begin{table*}[t]
\resizebox{\textwidth}{!}{
\begin{tabular}{lc@{~~~~}c@{~~}c@{~~}c@{~~}c@{~~}c@{~~}c@{~~}c@{~~}c@{~~}c@{~~}c@{~~}c@{~~}c@{~~}c@{~~}c@{~~}c@{~~}c@{~~}c@{~~}|c} \toprule
    & \rotatebox{90}{\small Make} \rotatebox{90}{\small Kimchi Rice}  
    & \rotatebox{90}{\small Pickle} \rotatebox{90}{\small Cucumber}  
    & \rotatebox{90}{\small Make Banana} \rotatebox{90}{\small Ice Cream}  
    & \rotatebox{90}{\small Grill} \rotatebox{90}{\small Steak}  
    & \rotatebox{90}{\small Jack Up } \rotatebox{90}{\small Car}  
    & \rotatebox{90}{\small Make } \rotatebox{90}{\small Jello Shots}  
    & \rotatebox{90}{\small Change } \rotatebox{90}{\small Tire}  
    & \rotatebox{90}{\small Make } \rotatebox{90}{\small Lemonade}  
    & \rotatebox{90}{\small Add Oil } \rotatebox{90}{\small to Car}  
    & \rotatebox{90}{\small Make } \rotatebox{90}{\small Latte}  
    & \rotatebox{90}{\small Build } \rotatebox{90}{\small Shelves}  
    & \rotatebox{90}{\small Make } \rotatebox{90}{\small Taco Salad}  
    & \rotatebox{90}{\small Make } \rotatebox{90}{\small French Toast}  
    & \rotatebox{90}{\small Make } \rotatebox{90}{\small Irish Coffee}  
    & \rotatebox{90}{\small Make } \rotatebox{90}{\small Strawberry Cake}  
    & \rotatebox{90}{\small Make } \rotatebox{90}{\small Pancakes}  
    & \rotatebox{90}{\small Make } \rotatebox{90}{\small Meringue}  
    & \rotatebox{90}{\small Make } \rotatebox{90}{\small Fish Curry}  
    & \rotatebox{90}{\small Average }
\\ \midrule
Fully-supervised upper-bound~\cite{zhukov2019crosstask}                           & 19.1          & 25.3          & 38.0          & 37.5          & 25.7          & 28.2          & 54.3          & 25.8          & 18.3         & 31.2          & 47.7          & 12.0          & 39.5          & 23.4          & 30.9          & 41.1          & 53.4          & 17.3          & 31.6 \\ \midrule
Alayrac \etal \cite{alayrac16unsupervised} & 15.6 & 10.6          & 7.5           & 14.2          & 9.3           & 11.8          & 17.3          & 13.1          & 6.4 & 12.9          & 27.2          & 9.2           & 15.7          & 8.6           & 16.3          & 13.0          & 23.2          & 7.4           & 13.3 \\
Zhukov \etal \cite{zhukov2019crosstask}                         & 13.3          & 18.0          & 23.4 & 23.1 & 16.9          & 16.5 & 30.7 & 21.6 & 4.6          & 19.5 & 35.3        & 10.0 & 32.3 & 13.8 & 29.5 & 37.6 & \textbf{43.0} & 13.3 & 22.4 \\ 
Ours trained on HowTo100M only                    & \textbf{33.5}           & \textbf{27.1}         & \textbf{36.6}           & \textbf{37.9}           & \textbf{24.1}           & \textbf{35.6} & \textbf{32.7}           & \textbf{35.1}           & \textbf{30.7}         & \textbf{28.5}           & \textbf{43.2}         & \textbf{19.8}           & \textbf{34.7}           & \textbf{33.6}           & \textbf{40.4}           & \textbf{41.6}          & 41.9           & \textbf{27.4}           & \textbf{33.6} \\
\bottomrule
\end{tabular}
}
\vspace{-2mm}
\caption{Step localization results on CrossTask~\cite{zhukov2019crosstask} instructional video dataset.}
\vspace{-3mm}
\label{table:cvpr19_results}
\end{table*}

\subsection{Comparison with state-of-the-art}

{\bf \noindent CrossTask.}
We compare our off-the-shelf embedding trained on HowTo100M against methods proposed by Alayrac~\etal~\cite{alayrac16unsupervised} and Zhukov~\etal~\cite{zhukov2019crosstask} which is the current state-of-the-art on CrossTask for weakly supervised methods. 
Note that Zhukov \etal~\cite{zhukov2019crosstask} have access to the ordered list of action labels at the task level and narrations are the only form of supervision during training.  
We also report the fully-supervised upper-bound from~\cite{zhukov2019crosstask} obtained with a model that has been trained on action segments with ground truth annotation.
The results are shown in Table~\ref{table:cvpr19_results}.
Our approach significantly outperforms the state-of-the-art, even though it has not been specifically designed for the task of step localization in videos. 
The improvement made by our method is consistent across all tasks (with the exception of \textit{Make Meringue}), showing that the trained model is not biased towards any specific domain.
The recall is above 30\% for most tasks with the significant improvement observed for the ``\textit{Add Oil to a Car}'' task (6.4\% to 30.7\% boost in recall).
Note that our method also outperforms the fully-supervised upper bound~\cite{zhukov2019crosstask} on average.
Thus, we conclude that training on a large amount of narrated videos is better than training a step localization model on a small but carefully annotated training set. 

\begin{table}[t]
  \setlength{\tabcolsep}{3pt}
    \centering  
        \resizebox{\linewidth}{!}{
      \begin{tabular}{@{}lccccc@{}}
      \toprule
      Method & Trainset & R@1 & R@5 & R@10  & Median R  \\
      \midrule
      Random & None & 0.03 & 0.15 & 0.3 & 1675 \\
      HGLMM FV CCA~\cite{klein15associating} & YouCook2 & 4.6 & 14.3 & 21.6 & 75   \\
      \midrule
      Ours & YouCook2 & 4.2 & 13.7 & 21.5 & 65 \\
      Ours & HowTo100M & 6.1 & 17.3 & 24.8 & 46 \\
      Ours & \small{\makecell{PT: HowTo100M \\ FT: YouCook2}} & \textbf{8.2} & \textbf{24.5} & \textbf{35.3} & \textbf{24} \\
      \bottomrule
    \end{tabular}
    }
   \vspace{-2mm}
 \caption{\small YouCook2 clip retrieval results. PT denotes: pre-trained, while FT denotes: fine-tuned.}
 \vspace{-2mm}
      \label{table:youcook-experiment}
\end{table}

{\bf \noindent YouCook2}~\cite{youcook2} does not provide an official benchmark nor any reported number for clip retrieval.
As a consequence, we have applied a state-of-the-art text-video embedding model from Klein~\etal~\cite{klein15associating} (HGLMM FV CCA) on YouCook2 using our features.
We also report results of our model trained on YouCook2 instead of HowTo100M in Table~\ref{table:youcook-experiment}.
First, we notice that our off-the-shelf model trained on HowTo100M significantly outperforms both the exact same model directly trained on YouCook2 and \cite{klein15associating}.
Furthermore, fine-tuning our model pre-trained on HowTo100M on YouCook2 results in a significant improvement of 13.7 \% in R@10 against \cite{klein15associating}.
In conclusion, we show that the off-the-shelf HowTo100M trained model can outperform state-of-the-art on this domain specific instructional video dataset.
Moreover, we demonstrate that our model can get further benefits from fine-tuning.

\begin{table}[t]
  \setlength{\tabcolsep}{3pt}
    \centering  
        \resizebox{\linewidth}{!}{
      \begin{tabular}{@{}lccccc@{}}
      \toprule
      Method & Trainset & R@1 & R@5 & R@10  & Median R  \\
      \midrule
      Random & None & 0.1 & 0.5 & 1.0 & 500 \\
      C+LSTM+SA+FC7~\cite{torabi16learning}     & MSR-VTT  & 4.2 & 12.9 & 19.9 & 55 \\
      VSE-LSTM~\cite{kiros14unifying} & MSR-VTT  & 3.8 & 12.7 & 17.1 & 66 \\ 
      SNUVL~\cite{yu16videocaptioning} & MSR-VTT & 3.5 & 15.9 & 23.8 & 44\\
      Kaufman \etal~\cite{kaufman17temporal}  & MSR-VTT  & 4.7 & 16.6 & 24.1 & 41\\
      CT-SAN~\cite{yu17endtoend}   & MSR-VTT  & 4.4 & 16.6 & 22.3 & 35\\
      JSFusion~\cite{yu18joint} & MSR-VTT & 10.2 & 31.2 & 43.2 & 13  \\
      \midrule
      Ours & HowTo100M & 7.5 & 21.2 & 29.6 & 38 \\
      Ours & MSR-VTT & 12.1 & 35.0 & 48.0 & 12 \\
      Ours & \small{\makecell{PT: HowTo100M \\ FT: MSR-VTT}} & \textbf{14.9} & \textbf{40.2} & \textbf{52.8} & \textbf{9} \\

      \bottomrule
    \end{tabular}
    }
  \vspace{-2mm}
 \caption{MSR-VTT clip retrieval results. PT denotes: pre-trained, while FT denotes: fine-tuned.}
  \vspace{-2mm}
      \label{table:msrvtt-experiment}
\end{table}

\begin{figure}[t]
  \centering
     \includegraphics[width=\columnwidth]{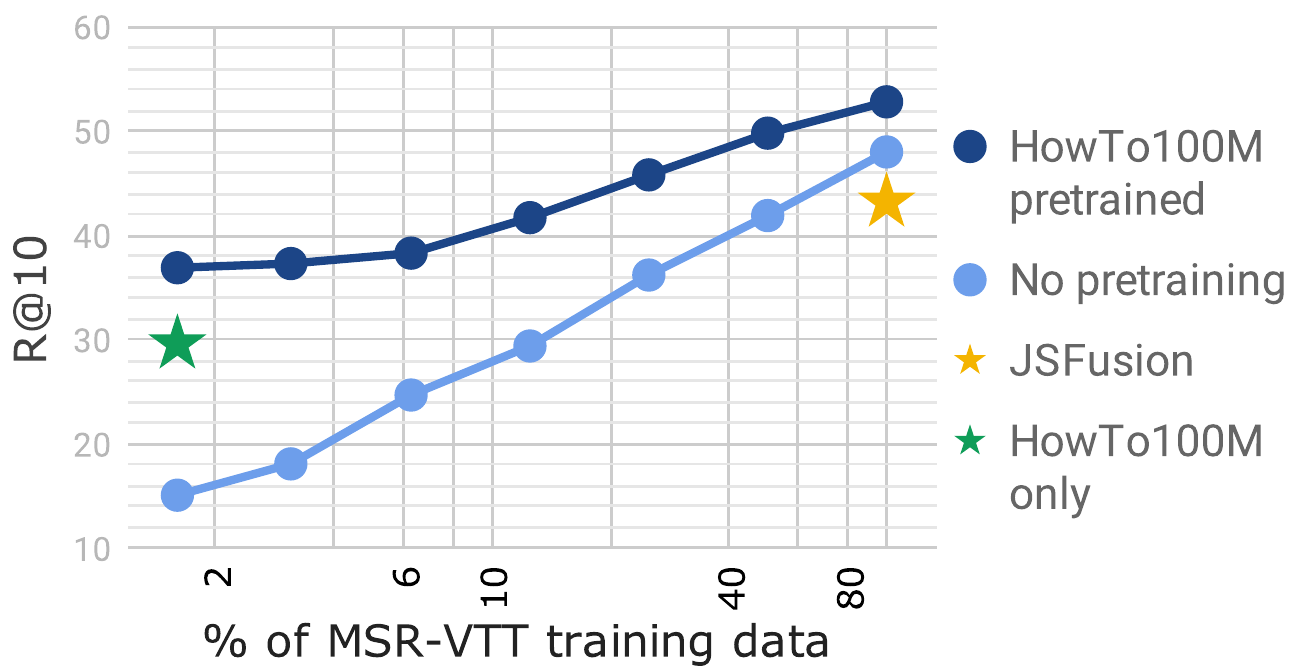}
  \vspace{-4mm}
  \caption{Evaluation of fine-tuning a HowTo100M pre-trained model with varying amounts of MSR-VTT supervision for text-to-video clip retrieval.}
  \label{fig:msrvtt_training}
\end{figure}

{\bf \noindent MSR-VTT.}
We compare our model trained on \textbf{(i)}~HowTo100M only, \textbf{(ii)}~MSR-VTT only and \textbf{(iii)}~pre-trained on HowTo100M and then fine-tuned on MSR-VTT against prior work that directly uses MSR-VTT for training (reproduced in~\cite{yu18joint}) in Table~\ref{table:msrvtt-experiment}.
Our off-the-shelf HowTo100M model outperforms~\cite{kaufman17temporal,kiros14unifying,torabi16learning,yu16videocaptioning,yu17endtoend} that are \textit{directly trained} on MSR-VTT.
Here again, after fine-tuning the HowTo100M pre-trained model on MSR-VTT, we observe a significant improvement over the state-of-the-art JSFusion~\cite{yu18joint} trained on MSR-VTT.
However, as opposed to instructional videos (CrossTask) and cooking videos (YouCook2), training our model directly on MSR-VTT performs better than our off-the-shelf model trained on HowTo100M. 
We believe this is due to MSR-VTT videos being generic Youtube videos that are different from the instructional or VLOG type of videos that dominate HowTo100M. 
In Figure~\ref{fig:msrvtt_training}, we also investigate the impact on performance at various amounts of supervision when fine-tuning our pre-trained model. 
It shows that state-of-the-art performance~\cite{yu18joint} can be attained with \emph{only} $20 \%$ of MSR-VTT samples.
This has great practical implications as comparable performance can be obtained using significantly reduced annotation.

\begin{table}[t]
  \setlength{\tabcolsep}{3pt}
    \centering  
        \resizebox{\linewidth}{!}{
      \begin{tabular}{@{}lccccc@{}}
      \toprule
      Method & Trainset & R@1 & R@5 & R@10  & Median R  \\
      \midrule
      Random & None & 0.1 & 0.5 & 1.0 & 500 \\
      C+LSTM+SA+FC7~\cite{torabi16learning}     & LSMDC  & 4.3 & 12.6 & 18.9 & 98 \\
      VSE-LSTM~\cite{kiros14unifying} & LSMDC  & 3.1 & 10.4 & 16.5 & 79 \\ 
      SNUVL~\cite{yu16videocaptioning} & LSMDC & 3.6 & 14.7 & 23.9 & 50\\
      Kaufman \etal~\cite{kaufman17temporal}  & LSMDC  & 4.7 & 15.9 & 23.4 & 64\\
      CT-SAN~\cite{yu17endtoend}   & LSMDC  & 4.5 & 14.1 & 20.9 & 67\\
      JSFusion~\cite{yu18joint} & LSMDC & \textbf{9.1} & \textbf{21.2} & \textbf{34.1} & \textbf{36}  \\
      \midrule
      Ours & HowTo100M & 4.0 & 9.8 & 14.0 & 137 \\
      Ours & LSMDC & \textbf{7.2} & 18.3 & 25.0 & 44 \\
      Ours & \small{\makecell{PT: HowTo100M \\ FT: LSMDC}} & 7.1 & \textbf{19.6} & \textbf{27.9} & \textbf{40} \\

      \bottomrule
    \end{tabular}
    }
  \vspace{-2mm}
 \caption{LSMDC clip retrieval results. PT denotes: pre-trained, while FT denotes: fine-tuned.}
  \vspace{-2mm}
      \label{table:lsmdc-experiment}
\end{table}

\begin{figure}[t]
  \centering
     \includegraphics[width=\columnwidth]{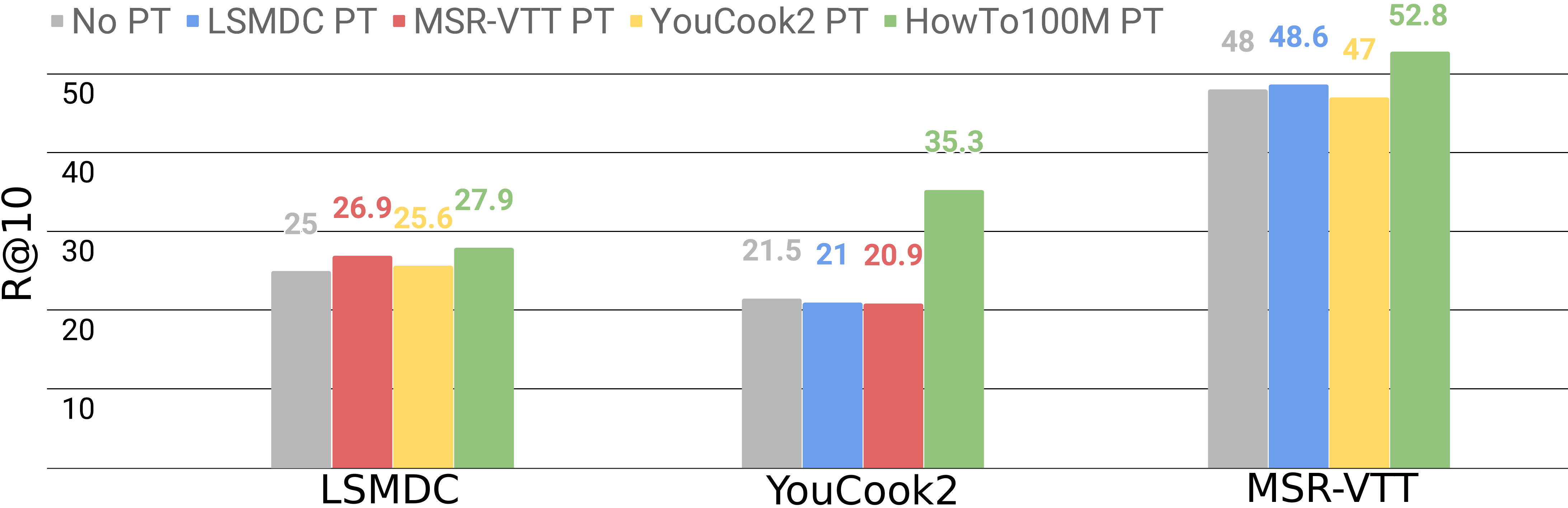}
  \vspace{-4mm}
  \caption{Results of clip retrieval by pre-training models on different datasets. Evaluation on LSMDC, YouCook2 and MSR-VTT.}
\label{fig:crossdataset}
\end{figure}

{\bf \noindent LSMDC.}
Finally, we compare to state-of-the-art on LSMDC in Table~\ref{table:lsmdc-experiment}.
This dataset is even more challenging as movie clips are quite distinct from HowTo100M videos.
We compare against several other prior works that have been reproduced in~\cite{yu18joint} and are trained directly on LSMDC.
Here again, we see that pre-training our model on HowTo100M and fine-tuning it on LSMDC also provides improvements upon a model directly trained on LSMDC.
This finding is interesting and shows that a HowTo100M pre-trained model can still be useful when fine-tuned on videos from a different domain.

\begin{figure}[t]
  \begin{center}
     \includegraphics[width=\columnwidth]{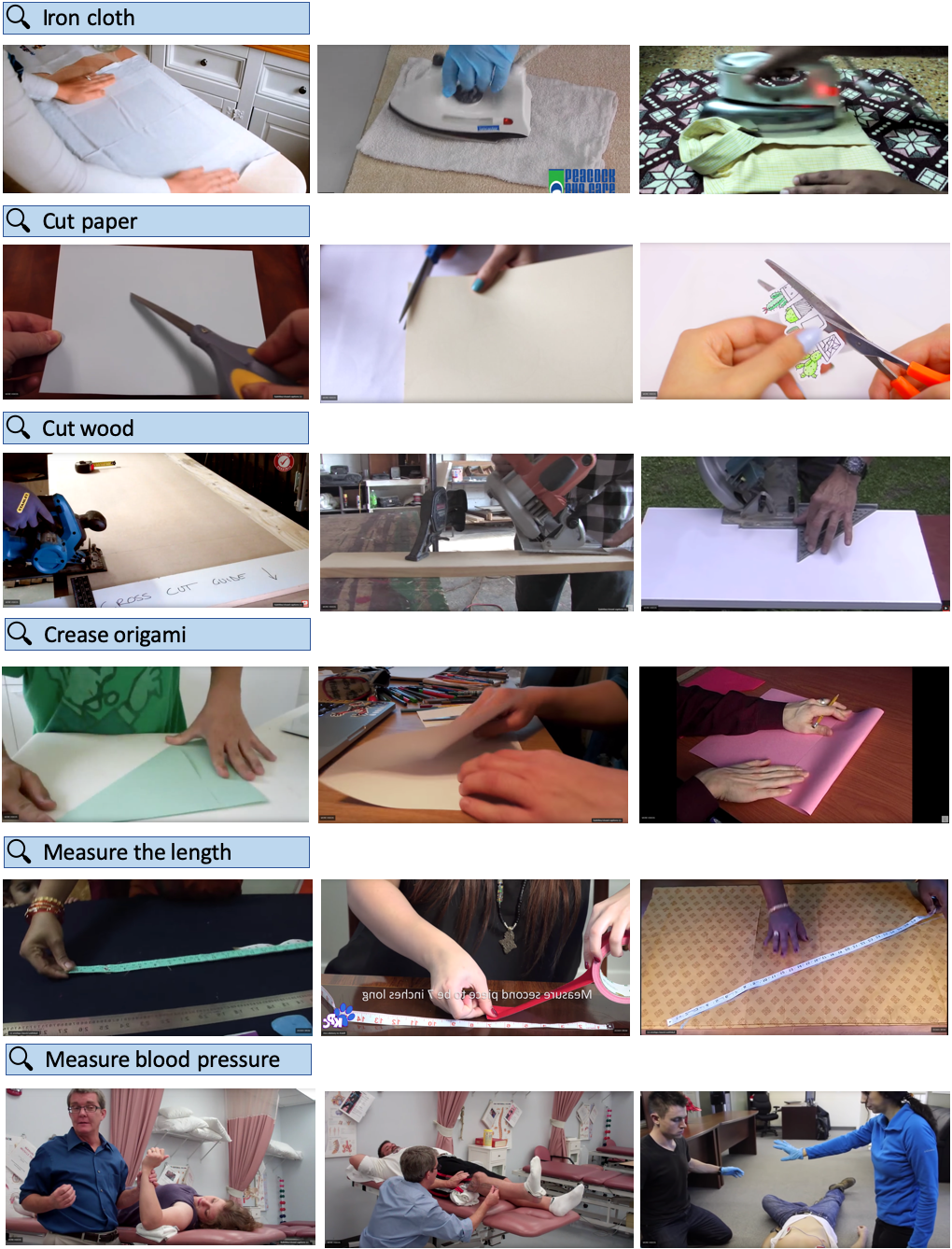}
\end{center}
\vspace{-0.7cm}
  \caption{Example video-clip retrieval results on HowTo100M using our trained joint embedding.}
\label{fig:qualitative}
\end{figure}

\subsection{Cross-dataset fine-tuning evaluation}
In this section, we evaluate the advantage of HowTo100M for pre-training compared to pre-training on other smaller datasets.
Figure~\ref{fig:crossdataset} shows evaluation on YouCook2, MSR-VTT and LSMDC clip retrieval (R@10) using no pre-training (No PT), using pre-training on YouCook2, MSR-VTT, LSMDC and HowTo100M datasets while fine-tuning to the target dataset.
For all evaluated datasets, pre-training on HowTo100M prior to fine-tuning on the target dataset consistently yields best results.

\subsection{Qualitative results}
Figure~\ref{fig:qualitative} illustrates examples of retrieved video clips from HowTo100M using our trained joint text-video embedding.
For example, our learned representation can correctly distinguish between queries \textit{Cut paper} and \textit{Cut wood}. 
A demo of the retrieval system is available online~\cite{iccv2019howto100m}.

\section{Conclusion}
We have introduced HowTo100M, a video dataset with more than 130M video clips, extracted from 1.2M narrated web videos of people performing complex visual tasks.
Our data collection method is fast, scalable and \textit{does not require any manual annotation}.
We use this dataset to learn a joint text-video embedding by leveraging more than 130M video clip-caption pairs.
We have shown through various experiments that our learned embedding can perform better compared to models trained on existing carefully annotated but smaller video description datasets.
The dataset, pre-trained models and code are available at \cite{iccv2019howto100m}.

\clearpage

\paragraph{Acknowledgements.}
The project was partially supported by 
Antoine Miech Google PhD fellowship, the MSR-Inria joint lab, the Louis Vuitton - ENS Chair on 
Artificial Intelligence, the ERC grant LEAP (No.\,336845), the CIFAR 
Learning in Machines\&Brains program, and the European Regional 
Development Fund under the project IMPACT (reg. no. 
CZ.02.1.01/0.0/0.0/15\_003/0000468).

{\small
\bibliographystyle{ieee}
\bibliography{master-biblio}
}
\clearpage

\clearpage
\appendix

\section*{Overview of Appendix}
We present additional details of our HowTo100M dataset in Appendix \ref{howto100mdetail}.
We also provide practical implementation details of our ranking loss in Appendix \ref{loss_detail} and analyze the sampling strategy for positive pair selection during training in Appendix \ref{positive_sampling}.

\section{Additional details of the HowTo100M dataset}  \label{howto100mdetail}
\begin{figure*}[t]
\begin{center}
   \includegraphics[width=\linewidth]{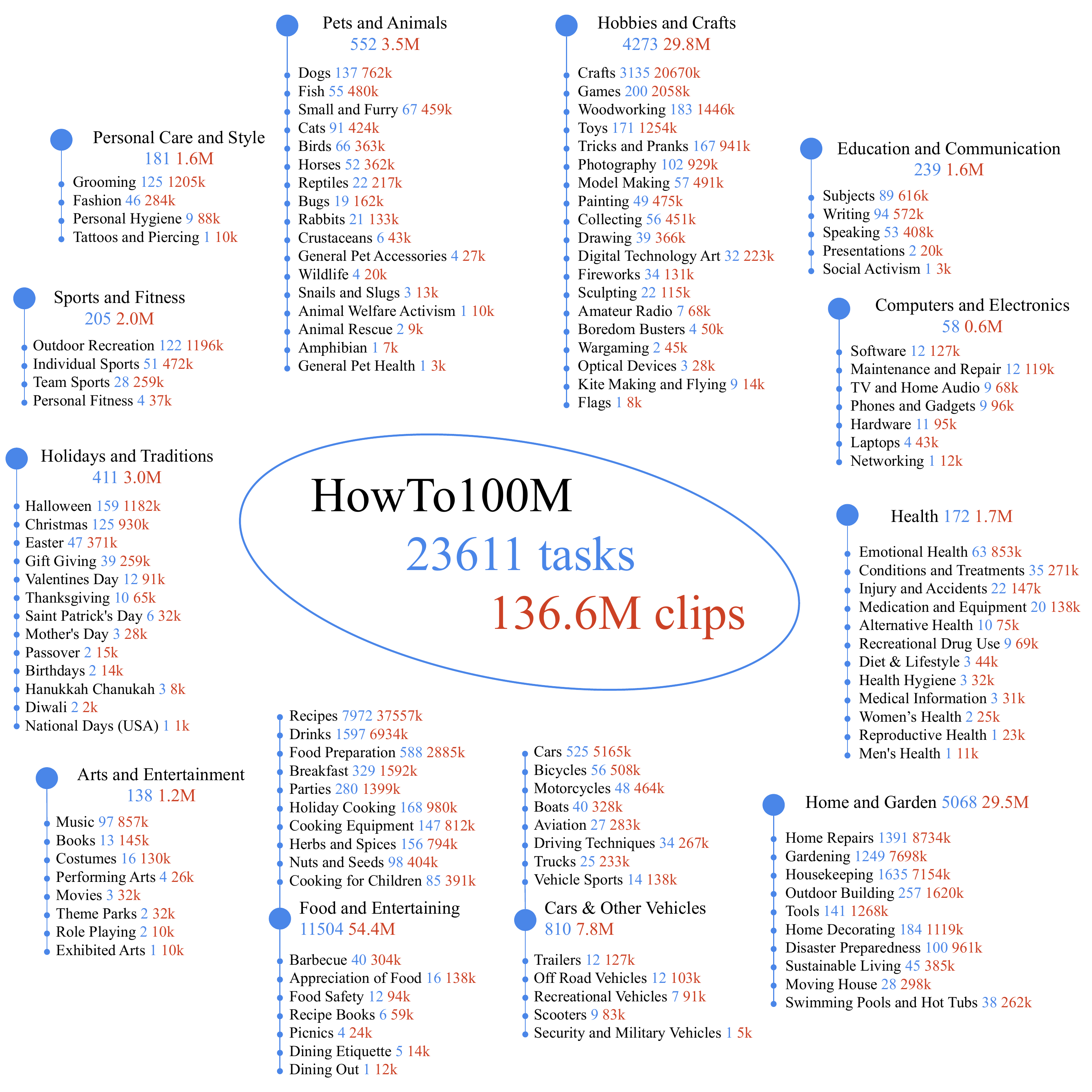}
   \caption{The first two levels of hierarchy of tasks in the HowTo100M dataset. Our dataset includes 12 categories from WikiHow containing 129 subcategories. For each (sub)category we show the total number of collected tasks and clips. This hierarchy of tasks in our dataset follows the WikiHow structure. Please recall that abstract tasks such as Choosing a gift or Meeting new friends, were not considered and were removed from the WikiHow hierarchy semi-automatically by verb analysis, as described in Section~3.1 of the main paper. As a result, the category tree is imbalanced. For example, the \textit{Dining Out} subcategory includes only one physical task (\textit{Fix a Shaky Table at a Restaurant}), while \textit{Recipes} subcategory from the same level of the hierarchy includes a large number of tasks and clips.}
   \label{fig:subcats}
\end{center}
\vspace{-0.2in}
\end{figure*}

Our HowTo100M dataset is based on the hierarchy of WikiHow\footnote{\url{https://www.wikihow.com/}} tasks.
The HowTo100M spans a total of 23,611 tasks. 
Here we visualize the first two levels of the WikiHow hierarchy -- the twelve categories and their subcategories, the number of underlying tasks and corresponding videos are illustrated in Figure~\ref{fig:subcats}.

\begin{figure*}[t]
\begin{center}
   \includegraphics[width=\linewidth]{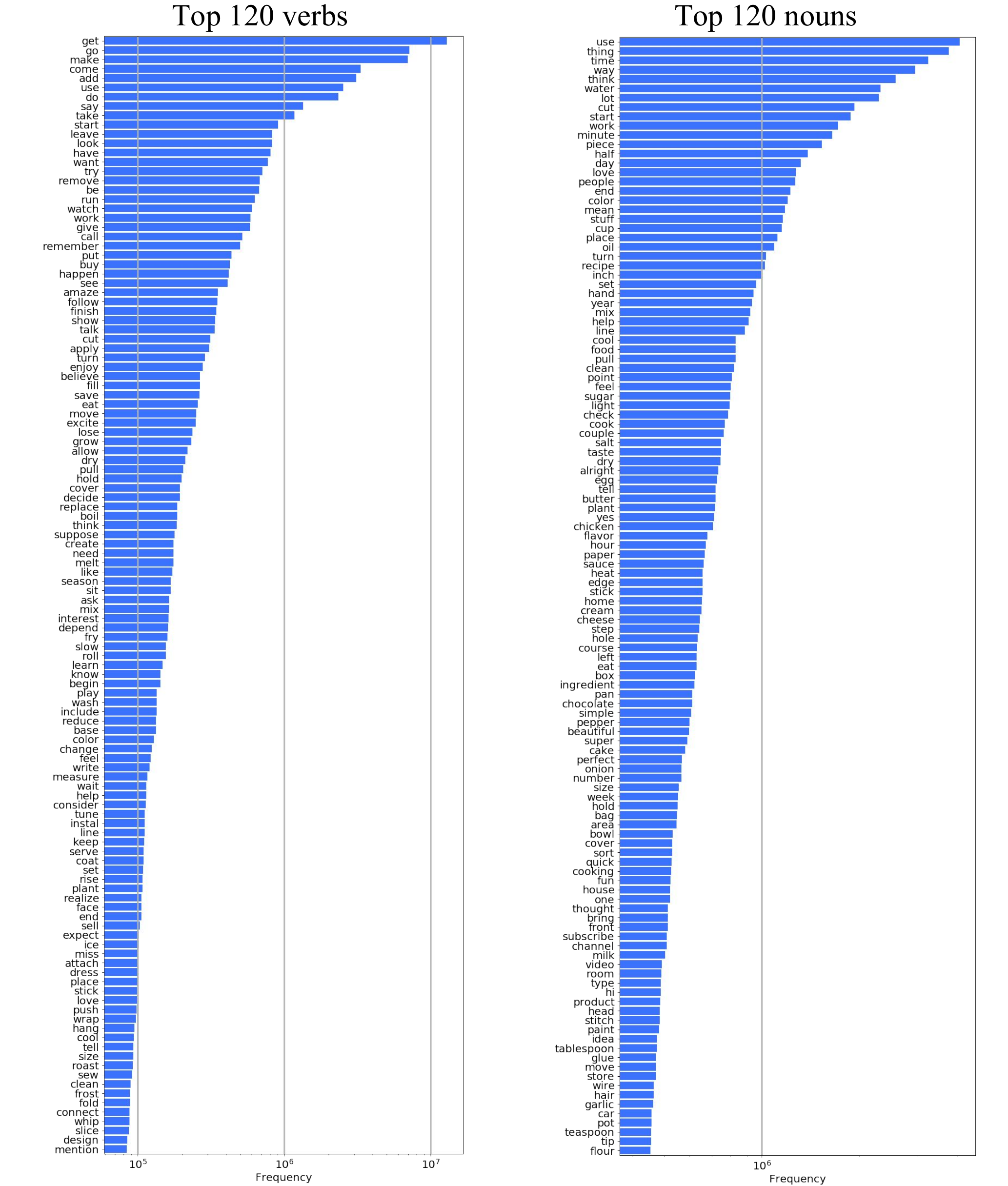}
   \caption{Frequencies of the top 120 most commonly occurring nouns and verbs in our dataset. Note that our dataset is biased towards physical actions, with verbs such as \textit{get}, \textit{go} and \textit{make} being the most frequent, while verbs, such as \textit{be}, \textit{know} and \textit{think} are less frequent than in common English. Top nouns show the dominant topics in our instructional videos. In particular, many cooking-related words, such as \textit{water}, \textit{oil} and \textit{sugar} occur with high frequency.}
   \label{fig:word_freqs}
\end{center}
\vspace{-0.2in}
\end{figure*}

HowTo100M comes with transcribed narrations which often describe the content of the videos.
Figure \ref{fig:word_freqs} shows frequencies of nouns and verbs in transcribed video narrations. 
We used the MaxEnt Treebank POS Tagger to obtain the nouns and verbs.
Please see the figure captions for additional analysis.

\begin{figure}[t]
\begin{center}
   \includegraphics[width=\columnwidth]{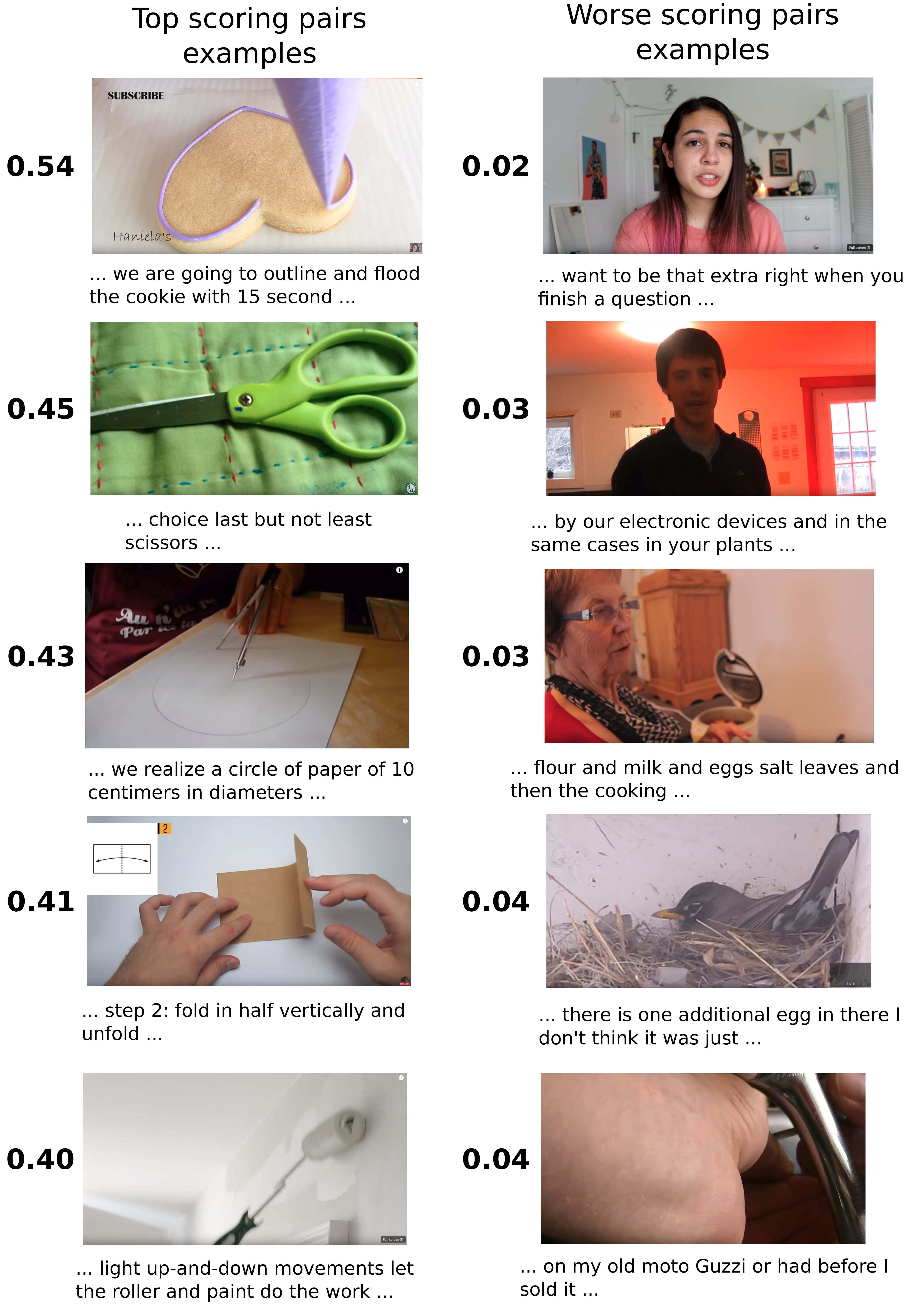}
   \caption{We illustrate examples of high and low scoring clip-caption pairs. 
   Examples from the left column show pairs where the caption visually describes what is seen in the corresponding video clip. On the other hand, low scoring pairs from the right column have captions that do not match visual content.}
   \label{fig:pairs_example}
\end{center}
\vspace{-0.2in}
\end{figure}

\section{Ranking loss implementation details}
\label{loss_detail}
In the main paper, we have defined our mini-batch ranking loss as:
\begin{align}
\sum_{i \in \mathcal{B}} \smashoperator[r]{\sum_{j \in \mathcal{N}(i)}}  \max(0, \delta + s_{i,j} - s_{i,i}) + \max(0, \delta + s_{j,i} - s_{i,i}).
\end{align}
We explain next how $\mathcal{N}(i)$ is constructed to improve computational efficiency.

At each training iteration, we first sample $v$ unique YouTube video ids.
We then sample with replacement a number $k$ of clip-caption pairs from each of these videos.
Therefore, we are left with a mini-batch containing $b=kv$ clip-caption pairs, with $v=32$ and $k=64$ in practice.
In order to not waste computation efforts, we use every sampled mini-batch pair as a negative anchor, \ie $\mathcal{N}(i) = \mathcal{B}\setminus \{i\}, \forall i$.

Doing so, the proportion of negative examples coming from the same video (\emph{intra-video}) is $\frac{k-1}{kv - 1}$ while the proportion of negatives from different videos (\emph{inter-video}) is $\frac{k(v - 1)}{kv - 1}$. 
A problem with this is that the ratio between \emph{intra} and \emph{inter} video negative examples depends on the number of unique videos sampled and the amount of clip-caption pairs collected per video (respectively $v$ and $k$). 
To address this, we follow ~\cite{hendricks17localizing} by re-weighting the inter-video and intra-video contributions inside the triplet loss. 
For example, in order to sample intra-video triplets with probability $p \in [0, 1]$ (and inter-video triplets with probability $1-p$), one can equivalently weight the intra-video triplet losses by: $\alpha = \frac{pk(v-1)}{(1-p)(k - 1)}$ (thus ensuring a ratio between intra-video and inter-video negative examples of $\frac{p}{1-p}$).
This allows us to fix the intra-video to inter-video negative sampling ratio regardless of $v$ and $k$.
Formally, we define the following weighting function:
\begin{equation}
  \alpha_{i,j} =\left\{
  \begin{array}{@{}ll@{}}
    \frac{pk(v-1)}{(1-p)(k - 1)} & \text{if}\ i \ \text{and} \ j \ \text{are from same video,} \\
    1, & \text{otherwise}.
  \end{array}\right.
\end{equation} 
We then use this weighing function to define the loss:
\label{loss}
\[
\sum_{i \in \mathcal{B}, j \in \mathcal{N}(i)} \alpha_{i,j}  \Big[ \max(0, \delta + s_{i,j} - s_{i,i}) + \max(0, \delta + s_{j,i} - s_{i,i})\Big].
\]

\section{Sampling strategy for positive pairs}
\label{positive_sampling}
As discussed in the main paper, narrations need not necessarily describe what is seen in the video.
As a consequence, some captions from HowTo100M do not correlate with their corresponding video clips (see Figure \ref{fig:pairs_example}).
To deal with this noisy data, we tried a sampling strategy for positive pairs that 
aims to discard non-relevant video-caption pairs during training.
Inspired by multiple instance learning, our idea is to select a subset of top scoring clip-caption  training pairs within each video.

In particular, given a video with $N$ video clip-caption pairs $\{(V_i, C_i)\}_{i \in [1, N]}$, 
we first compute the similarity scores of all the $N$ pairs: $s(V_i, C_i)$ using the current model parameters.
We then use a pre-defined max-pool rate $r \in [0, 1]$ of the highest scoring positive training pairs $\{(V_i, C_i)\}_{i \in [1, N]}$ within each video.
For example, at $r = 0.5$ we retain the high scoring half of all $N$ pairs for training.

Table \ref{table:positive-sampling-experiment} shows results of our positive sampling strategy when varying the max pool rate $r$ with evaluation on video clip retrieval.
For example, $r = 1.0$ means that no sampling strategy is applied as we keep all $N$ pairs as potential candidates.
Interestingly, in our case, carefully selecting the positive pairs does not improve our model as the best results are obtained with $r = 1.0$. 
Note that decreasing the max pool rate also decreases the number of triplet losses computed within a mini-batch by the same rate.
To show that the number of triplet losses computed for each mini-batch does not impact the overall performance, we have performed a sanity check experiment in Table \ref{table:positive-sampling-check} in which we also replaced the max pool sampling by random sampling of pairs for $r =0.5$. The results with random sampling at $r=0.5$ are very similar to the results obtained with no max pool sampling (r=1.0) as shown in Table \ref{table:positive-sampling-experiment}, which confirms our finding that our model is relatively robust to the noisy positive pairs. We think this could be attributed to the fact our model is shallow and is trained on a large amount of data.

\begin{table}[t]
  \setlength{\tabcolsep}{3pt}
    \centering  
      \begin{tabular}{@{}lccccc@{}}
      \toprule
      Max pool rate (r) & M (R@10) & L (R@10) & Y (R@10)  \\
      \midrule
      0.2  & 21.9  & 13.9 & 19.7 \\
      0.5  & 25.2 & 12.6  & 23.5 \\
      0.9  & 27.3 & 12.6 & 23.9  \\
      1.0  (no max pool) &  \textbf{29.6} & \textbf{14.0} & \textbf{24.8} \\
      \bottomrule
    \end{tabular}
\vspace{-2mm}
 \caption{Study of positive pair sampling. When max pool rate r is below 1.0 only the proportion r of top scoring clip-caption pairs are used for learning.   We report R@10 retrieval results from M: MSR-VTT, L: LSMDC, Y: YouCook2.}
 \vspace{-4mm}
      \label{table:positive-sampling-experiment}
\end{table}

\begin{table}[t]
  \setlength{\tabcolsep}{3pt}
    \centering  
      \begin{tabular}{@{}lccccc@{}}
      \toprule
      MP rate & RS rate &M (R@10) & L (R@10) & Y (R@10)  \\
      \midrule
      1.0  & 0.5 & \textbf{28.8}  & \textbf{14.3} & \textbf{24.2} \\
      0.5  & 1.0 & 25.2 & 12.6  & 23.5 \\
      \bottomrule
    \end{tabular}
\vspace{-2mm}
 \caption{Study of Random Sampling (RS) vs. Max Pool (MP) sampling of positive clip-caption pairs. We report R@10 retrieval results from M: MSR-VTT, L: LSMDC, Y: YouCook2.}
 \vspace{-4mm}
      \label{table:positive-sampling-check}
\end{table}

\end{document}